%% file: main.tex
\documentclass[10pt,twocolumn,letterpaper]{article}

\usepackage[pagenumbers]{cvpr} % To force page numbers, e.g. for an arXiv version
\usepackage{algorithm}
\usepackage{algpseudocode}

\input{preamble}

\definecolor{cvprblue}{rgb}{0.21,0.49,0.74}
\usepackage[pagebackref,breaklinks,colorlinks,citecolor=cvprblue]{hyperref}

\def\paperID{40} % *** Enter the Paper ID here
\def\confName{CVPR}
\def\confYear{2024}

\title{Minimizing Embedding Distortion for Robust Out-of-Distribution Performance}

\author{Tom Shaked\thanks{Equal contribution.} \qquad Yuval Goldman\footnotemark[1] \qquad Oran Shayer\\
AppsFlyer\\
{\tt\small \{tom.shaked,yuval.goldman,oran.shayer\}@appsflyer.com}
}

\begin{document}
\maketitle
\input{sec/0_abstract}    
\input{sec/1_intro}

\input{sec/2_relatedwork}
% \input{sec/3_background}
\input{sec/4_approach}
\input{sec/5_experminents}
\input{sec/6_conclusions}
{
    \small
    \bibliographystyle{ieeenat_fullname}
    \bibliography{main}
}

\input{sec/X_suppl}

\end{document}

%% file: preamble.tex
\usepackage[dvipsnames]{xcolor}

%% file: sec/0_abstract.tex
\begin{abstract}

Foundational models, trained on vast and diverse datasets, have demonstrated remarkable capabilities in generalizing across different domains and distributions for various zero-shot tasks. Our work addresses the challenge of retaining these powerful generalization capabilities when adapting foundational models to specific downstream tasks through fine-tuning. To this end, we introduce a novel approach we call "similarity loss", which can be incorporated into the fine-tuning process of any task. By minimizing the distortion of fine-tuned embeddings from the pre-trained embeddings, our method strikes a balance between task-specific adaptation and preserving broad generalization abilities. We evaluate our approach on two diverse tasks: image classification on satellite imagery and face recognition, focusing on open-class and domain shift scenarios to assess out-of-distribution (OOD) performance. We demonstrate that this approach significantly improves OOD performance while maintaining strong in-distribution (ID) performance.

% \renewcommand{\thefootnote}{}
% \footnotetext{This is an unnumbered footnote.}

\footnotetext[1]{Accepted to ECCV 2024 Workshop.}

\end{abstract}

%% file: sec/1_intro.tex
\section{Introduction}
\label{sec:intro}

% One of the primary challenges in the field of machine learning is achieving strong generalization capabilities \cite{kawaguchi2017generalization, zhang2021understanding}. In real-world applications, our ultimate goal is to train models that can perform as effectively on unseen data as they do on the distribution of data used for training. Indeed, in many practical scenarios, the ability to generalize at an open-world scale is crucial. For instance, in autonomous driving systems \cite{yurtsever2020survey, caesar2020nuscenes}, we expect the models to perform reliably under any given weather condition, even those that are impossible to predict or simulate during training. Similarly, in robotics applications \cite{kober2013reinforcement}, we want robots to make accurate decisions in any environment, and in aerial imagery analysis \cite{liu2024remoteclip, pritt2017satellite}, models need to generalize across diverse geographic regions and environmental conditions.
One of the primary challenge in machine learning is achieving strong generalization capabilities \cite{kawaguchi2017generalization, zhang2021understanding}, particularly in real-world applications where models must handle unpredictable scenarios. This is crucial in diverse fields such as autonomous driving \cite{yurtsever2020survey, caesar2020nuscenes}, robotics \cite{kober2013reinforcement}, and aerial imagery analysis \cite{liu2024remoteclip, pritt2017satellite}, where systems must adapt to varied conditions. The need to handle this long-tail of possible scenarios \cite{wang2017learning} demands vast, comprehensive datasets. However, creating such datasets is often infeasible, as many scenarios are too rare or unpredictable to be adequately represented in training data, posing a significant challenge for model development and deployment.

\begin{figure}[t]
\centering
\includegraphics[width=8cm, height=2.7cm]{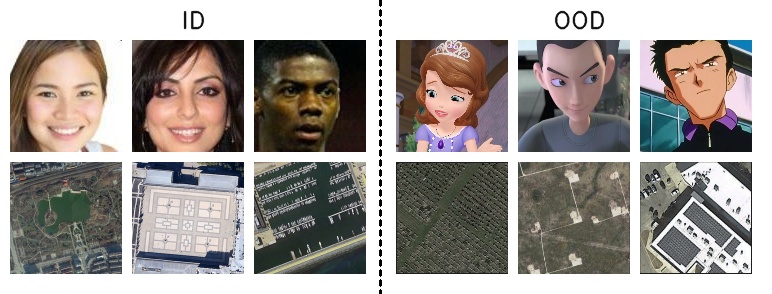}

\caption{Images sampled from the ID and OOD datasets for face recognition (top - domain shift) and image classification (bottom - unseen classes, for example the rightmost image is labeled "solar panel", not present in train set).}
\label{fig:id_ood_examples} 
\end{figure}

% The emergence of foundational models \cite{singh2022flava, clip} has shown promise in addressing these generalization challenges. Trained on massive and diverse datasets \cite{goodson2023huggyflan}, these models have demonstrated remarkable zero-shot and few-shot learning capabilities across various domains. However, when adapting them to specific downstream tasks through fine-tuning \cite{touvron2023llama}, we often encounter a trade-off between task-specific performance and the preservation of their broad generalization capabilities. Fine-tuning can lead to overfitting on the target distribution, potentially compromising the model's ability to handle out-of-distribution (OOD) scenarios effectively. \textcolor{red}{A lot of citations in this paragraph don't make sense. Please switch them to others @yuval}
Foundational models \cite{clip,kirillov2023segment}, trained on massive and diverse datasets \cite{goodson2023huggyflan}, have shown promise in addressing these generalization challenges with remarkable zero-shot and few-shot learning capabilities. However, adapting them to specific downstream tasks through fine-tuning \cite{mukhoti2023fine,Kumar2022FineTuningDistorts} often leads to a trade-off between task-specific performance and preservation of broad generalization capabilities, potentially compromising the model's ability to handle OOD scenarios.

In this paper, we take a step towards "enjoying both worlds" - tuning the embedding space for our specific task while leveraging the powerful generalized embeddings from pretrained models. We propose a novel "similarity loss" approach that can be incorporated into the fine-tuning process of any task. Our method aims to minimize the distortion of fine-tuned embeddings from their pre-trained counterparts, striking a balance between task-specific adaptation and the preservation of broad generalization abilities. This approach allows us to train on downstream tasks without discarding the strong generalizability capabilities of foundation models.

We evaluate our approach on two challenging tasks: image classification on satellite imagery and face recognition. These domains are well suited for assessing OOD performance due to their inherent variability and the potential for significant distribution shifts (fig. \ref{fig:id_ood_examples}). Our experiments focus on open-class and domain shift scenarios, providing a comprehensive assessment of our method's effectiveness in improving OOD performance.
We demonstrate that our method significantly improves OOD performance while incurring only a small reduction in ID performance across different tasks.
We provide extensive experiments and analysis, showcasing the effectiveness of our approach in real-world scenarios with potential distribution shifts.

% The rest of this paper is organized as follows: section \ref{sec:relwork} discusses related work, section \ref{sec:approach} details our proposed approach, section \ref{sec:experiments} describes our experimental setup and results, and section \ref{sec:conclusion} concludes with a discussion of our findings and future directions.

%% file: sec/2_relatedwork.tex
\section{Related Work}
\label{sec:relwork}

Adapting pre-trained foundation models to downstream tasks while maintaining robustness is an ongoing challenge. \cite{Kumar2022FineTuningDistorts} showed that fine-tuning can distort pretrained features and underperform linear probing on OOD data, proposing LP-FT as a solution. Several approaches followed: WiSE-FT \cite{Wortsman2021} used weight-space ensembling, CLIPood \cite{clipood} employed margin metric softmax and Beta moving average, and FLYP \cite{flyp} continued using the contrastive loss from pretraining. These methods have shown varying success in improving OOD performance while maintaining in-distribution accuracy.

Other works have explored lightweight fine-tuning approaches to preserve pretrained features, such as prompt-based methods \cite{Wang2021LearningTP}. Concurrent work by \cite{wortsman2022} showed that ensembling the weights of zero-shot and fine-tuned models can help balance ID and OOD performance. These studies highlight the importance of carefully considering the fine-tuning process to maintain the generalization capabilities of foundation models like CLIP \cite{clip}.

% In the context of satellite imagery, we use RemoteCLIP \cite{liu2024remoteclip} as our benchmark. This model, a fine-tuned variant of CLIP, represents the current state-of-the-art in this domain. For face recognition, we employ ArcFace \cite{deng2019arcface} as our baseline, given its widespread adoption and strong performance.

%% file: sec/4_approach.tex
\section{Approach}\label{sec:approach}
\subsection{Similarity Loss}
One of the key advantages of leveraging pre-trained foundation models for downstream tasks is the fact that these models are usually trained on massive amounts of data, covering a diverse set of domains and achieving impressive generalization on them.
This broad generalization is achieved in terms of both the coverage of open-world vocabulary, and a wide range of data domains.
Our goal is to perform fine-tuning of such foundation models on a downstream task while preserving these generalization capabilities.

Our key finding is that while these models achieve impressive zero-shot generalization across many domains and tasks, fine-tuning on a specific downstream task distorts the embedding space, fitting it to the dataset's domain and hurting generalization to other domains. We therefore argue that additional constraints during fine-tuning are necessary to prevent this distortion and preserve the pre-trained model's generalization capabilities.
At the heart of our approach lies the similarity loss. It is a simple constraint which can also be viewed as a form of regularization, that can be clipped onto any loss function and it is task independent. It is formalized as follows:

\begin{equation}\label{eq:sim_loss}
    L_{sim}(x) = \|f_{\theta}(x) - f_{\theta_0}(x)\|_2^2
\end{equation}

where x is the input, $f_{\theta_0}$ is the original pre-trained model, kept frozen through the entire fine-tuning phase, and $f_{\theta}$ is the model we train, initialized with the pre-trained weights from $f_{\theta_0}(x)$.
% In the next subsections (\ref{approach_classificaiton, approach_facerec}) we show how we easily combine it in the training procedure of two different tasks.

The similarity constraint can be weighted in the total loss by a coefficient $\alpha$, which balances task-specific specialization with the preservation of the pre-trained model's semantic properties. Notice that when $\alpha = 0$, standard unconstrained fine-tuning occurs, while $\alpha \to \infty$ essentially maintains the original pre-trained model. We can then control the value of $\alpha$ in training according to how we want to balance between ID and OOD performance.

In this paper, we focus on the CLIP model \cite{clip, openclip} to demonstrate our approach, but this can be applied to different foundation models, depending on the task at hand.

\subsection{Similarity Loss for Image Classification}\label{approach_classificaiton}

\textbf{Framework.} We utilize CLIP in a manner similar to that described in \cite{flyp} and \cite{clipood}. Specifically, we fine-tune the model on the downstream task in a contrastive manner, without introducing an additional linear classification layer. During training and inference, we convert class labels into a caption format, following the template "a photo of a \{class\}".

\textbf{Loss Function.} We imply the following loss:
\begin{equation}\label{eq:classification_loss}
    \mathcal{L} = L_\text{clip} \left(I_{1:B},T_{1:B}\right) + \alpha \cdot \frac{1}{B} \sum_{i=1}^B L_\text{sim} \left(I_i \right),
\end{equation}
where $B$ is the batch size, and $I,T$ are the image-text pairs.

This loss is composed of the standard CLIP loss \cite{clip} referred as $L_{clip}$ and our similarity loss from \ref{eq:sim_loss}. 
% $\alpha$ is a hyperparameter controlling the strength of the similarity constraint, providing a trade-off between the specialization on the specific downstream task and domain of the dataset, and the generalization of the original pretrained embedding space.

For this task, we imply the similarity loss over the vision encoder embeddings. In our experiments, we found that extending this similarity constraint to the text encoder as well did not yield additional improvements. We hypothesize that this is because CLIP loss aligns the vision and text embeddings, hence the similarity loss over the vision encoder indirectly constrains the text encoder as well.

% \textcolor{blue}{The loss combines CLIP loss \cite{clip} referred as $L_{clip}$ and our similarity loss \ref{eq:sim_loss}. $\alpha$ balances task-specific specialization and generalization. We apply similarity loss to vision encoder embeddings only. Extending this constraint to the text encoder didn't yield improvements, likely because CLIP loss already aligns vision and text embeddings, indirectly constraining the text encoder through the vision encoder.}

% \begin{equation}
% \begin{aligned}
% \mathcal{L}_\text{con} =
% - \frac{1}{2N} \sum_{i=1}^N \log \frac{\exp(\text{sim}(i_i, t_i) / \tau)}{\sum_{j=1}^N \exp(\text{sim}(i_i, t_j) / \tau)} \\
% - \frac{1}{2N} \sum_{i=1}^N \log \frac{\exp(\text{sim}(t_i, i_i) / \tau)}{\sum_{j=1}^N \exp(\text{sim}(t_i, i_j) / \tau)}
% \end{aligned}
% \end{equation}

% Where \(N\) represents the number of image-text pairs in a batch, \(i_i\), \(t_i\) are the image and text embedding respectively, \(\text{sim}(\cdot, \cdot)\) is the cosine similarity function and a \(\tau\) is the temperature parameter.

\subsection{Similarity Loss for Face Recognition}\label{approach_facerec}

\textbf{Framework.} For the face recognition task, we leverage CLIP as the pretrained model to fine-tune.
While CLIP's architecture and pre-training data are not specifically tailored for face recognition, as opposed to common approaches in this field \cite{deng2019arcface, huang2020curricularface, kim2020groupface, li23bionet} that tailor the architecture for this task, we see value in examining our method in this context.
Although there is currently no foundation model specifically designed for facial data, and we acknowledge that our in-distribution performance will not match recent state-of-the-art models, we believe there is room to show the benefits of this approach for this task, and that the results we will show in sec. \ref{sec:experiments} on extreme domain shifts will serve as an additional validation source of our claim.

Lacking paired text or semantic class labels, we use only the vision encoder, tuning it to capture discriminative facial features. Unlike standard CLIP training, which uses image-text pairs for contrastive loss, we train on image-image pairs. For each example in the batch we sample a pair of images $U_i, V_i$ that match the same identity, and will serve as the positive pair.

\textbf{Loss function.} Following the suggested framework, our loss function takes the following form:
\begin{equation}\label{eq:face_rec_loss}
    \mathcal{L} = L_\text{clip} \left(U_{1:B},V_{1:B}\right) + \alpha \cdot \frac{1}{B} \sum_{i=1}^B \left(L_\text{sim} \left(U_i \right) + L_\text{sim} \left(V_i \right)\right)
\end{equation}
We imply the similarity loss over both facial images in each pair, $U_i$ and $V_i$.

\textbf{Training procedure} is outlined in algorithm \ref{alg:face_recognition}. The contrastive nature of the training process necessitates careful batch construction. We build each batch by sampling face images such that all identities within the batch are distinct. The overall algorithm for face recognition training.

\begin{algorithm}
\caption{Face Recognition Training Procedure}
\label{alg:face_recognition}
\begin{algorithmic}[1]
\Require Pre-trained CLIP model $f_{\theta_0}$
\Require Facial images dataset $\mathcal{D}$ with $N$ unique identities
\State Initialize the fine-tuned model $\theta \leftarrow \theta_0$
\For{$k=1$ \textbf{to} $K$}
    \State Create batch of size $B$ by sampling $(U,V)$ image pairs from $B$ unique identities
    \State Calculate $\mathcal{L}$ as in equation \ref{eq:face_rec_loss} % Assuming the equation number is 1
    \State Update model parameters $\theta \leftarrow \theta - \eta \nabla_{\theta}\mathcal{L}$
\EndFor
\end{algorithmic}
\end{algorithm}

%% file: sec/5_experminents.tex
\section{Experiments}\label{sec:experiments}

\begin{figure}
  \centering
  \includegraphics[width=0.47\textwidth]{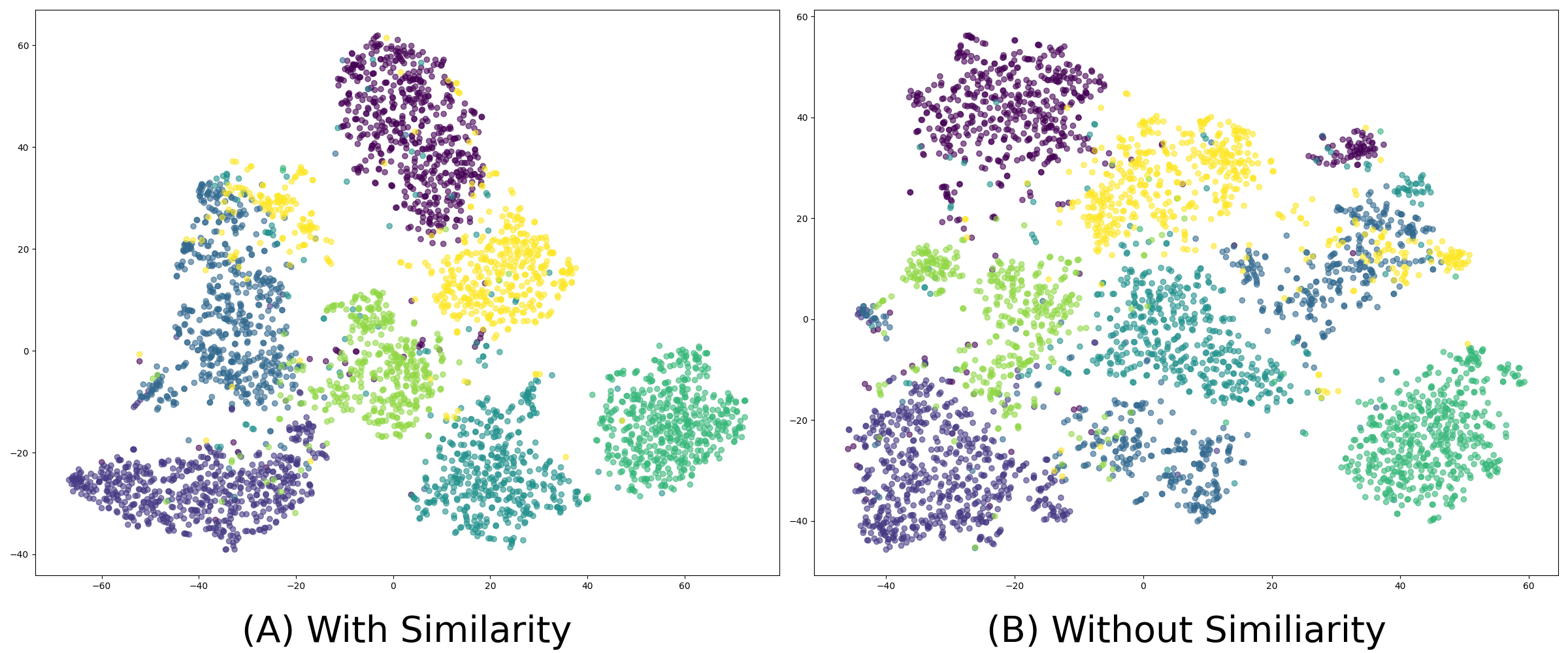}
\caption{Image embeddings of the OOD EuroSAT dataset, color-coded by class.
(A) Model trained with similarity loss (Avg. cluster variance: 1.87e-04)
(B) Model trained without similarity loss (Avg. cluster variance: 4.00e-04)}
\end{figure}\label{fig:embeddings}

\begin{table*}[t!]
\scriptsize
\centering
\begin{tabular}{lcccc|ccc}
\toprule
                                   & \multicolumn{4}{c}{RSICD (ID)}    &  EuroSAT (OOD)  &  RS-ICB128 (OOD) &  PatternNet (OOD) \\ \midrule
\multicolumn{1}{l}{Method}         & RET@1 & RET@5 & RET@10 & Mean RET &   Accuracy    &    Accuracy     &     Accuracy     \\ \midrule
RemoteCLIP \cite{liu2024remoteclip}                         & \textbf{17.02} & \textbf{37.97} & \textbf{51.51} & \textbf{35.26} & 35.9 & 24.18 & 57.81 \\
CLIP pre-trained \cite{openclip}                   & 4.75  &  18.3 & 28.6   & 17.2     & 43.26         & 26.62           & \textbf{64.3}    \\
Ours baseline (no similarity loss) & 12.5  & 31.9  & 51.0   & 32.3     & 26.81         & 19.3            & 38.49            \\
Ours + similarity loss             & 11.4  & 35.1  & 51.3   & 32.6     & \textbf{51.2} & \textbf{34.83}  & 63.41            \\ \midrule 
\end{tabular}
\caption{Evaluation results on ID and OOD datasets for the satellite image task.}
\label{tab:image_classification}
\end{table*}

\begin{table*}[t!]
\scriptsize
\centering
\begin{tabular}{lccc|cc}
\toprule
                                   & \multicolumn{3}{c}{IJBC (ID)}                    & \multicolumn{2}{c}{iCartoonFace (OOD)} \\ \midrule
\multicolumn{1}{l}{Method}         & TAR @ 1e-6     & TAR @ 1e-5     & TAR @ 1e-4     & TAR @ 0.01         & TAR @ 0.05        \\ \midrule
ArcFace \cite{deng2019arcface}                            & \textbf{89.97} & \textbf{94.34} & \textbf{96.18} & 4.85               & 13.04             \\
CLIP pre-trained \cite{openclip}                    & 5.11           & 13.58          & 27.44          & 32.97              & 56.55             \\
Ours - CLIP, contrastive, no similarity loss & 18.49          & 34.55          & 55.21          & 20.07              & 37.85             \\
Ours - CLIP, contrastive, with similarity loss             & 13.12          & 23.64          & 40.96          & \textbf{39.67}     & \textbf{62.29}    \\
Ours - CLIP, ArcLoss, no similarity loss                     & 39.01          & 47.14          & 60.14          & 10.09              & 25.13             \\
Ours - CLIP, ArcLoss, with similarity loss    & 7.35           & 18.94          & 34.7           & 36.14              & 58.69             \\ \midrule
\end{tabular}

\caption{Evaluation results on ID and OOD datasets for the face recognition task.}
\label{tab:face_recognition}
\end{table*}

To evaluate the effectiveness of our similarity loss approach, we conduct experiments across two distinct OOD scenarios, as mentioned in sec. \ref{sec:intro} and detailed in \cite{clipood}: unseen-classes in the test set and extreme domain shift.%These diverse OOD scenarios demonstrate the broad applicability of our method in enhancing model generalization across different tasks and distribution shifts.

All of our models were initialized with weights of ViT-B/32-laion2b pre-trained model from OpenCLIP \cite{openclip}. Further implementation details are provided in the supplement.

\subsection{Image Classification}
For the task of image classification we focus on the satellite imagery domain.
We examine OOD performance by testing on a distribution shift with a possible open-class scenario, where the test set might contain classes not seen in the fine-tuning train set.

\subsubsection{Datasets}
\textbf{RSICD} \cite{lu2017exploring} is comprised of 10,921 remote sensing images collected from Google Earth, Baidu Map, MapABC, and Tianditu. The dataset is split into 8,734 training images and 1,093 test images.
Each image is paired with 5 text captions, together they comprise 54,605 image-text pairs.
This dataset will serve as our ID dataset.

\textbf{EuroSAT} \cite{helber2019eurosat} contains 27,000 geo-referenced samples of Sentinel-2 satellite imagery, labeled into 10 land use and land cover classes. \textbf{RS-ICB128} \cite{li2020rsi} consists of 36,707 128x128 pixel images of 45 scene categories, sourced globally to represent China's land use classification standards. It's valuable for land cover and land use analysis. \textbf{PatternNet} \cite{zhou2018patternnet} includes 30,400 high-resolution (256x256 pixel) images across 38 classes, with 800 images per class. These images are sourced from Google Earth and Map API. These datasets will serve as our OOD datasets, showcasing a distribution shift as they are gathered from different data sources than RSICD, and present an open-class case study as well, where some classes and content differ significantly from our ID dataset (fig. \ref{fig:id_ood_examples}).

\subsubsection{Evaluation protocol}

% % ======

For the ID test set (RSICD), we utilize the Retrieval@K (RET@K) metric, where K = 1, 5, and 10. This metric is the common metric for image-caption paired data \cite{liu2024remoteclip}. RET@1 effectively measures the model's accuracy in exact caption retrieval, while RET@5 and RET@10 provide insights into broader retrieval performance. For OOD datasets, we employ the standard classification accuracy metric, which is more suitable for the multi-class classification nature of these datasets.

\subsubsection{Results}

We used RemoteCLIP \cite{liu2024remoteclip} as our primary comparison baseline, as it has shown strong performance on this task and trained in a similar manner. We began by evaluating the pretrained CLIP model without fine-tuning, followed by our fine-tuned model with and without the custom similarity component.

As shown in Table \ref{tab:image_classification}, our custom similarity component significantly improved performance on the OOD datasets, as reflected in accuracy metrics, while introducing only a minor decline in ID performance. Figure \ref{fig:embeddings} illustrates the embeddings from the OOD dataset EuroSAT, where the model that trained with our similarity loss exhibits substantially lower average cluster variance compared to the model trained without it.

While we observed a slight reduction in ID performance compared to the results reported by RemoteCLIP, our model consistently outperformed on all OOD datasets. Notably, for EuroSAT and RS-ICB128, our model not only exceeded RemoteCLIP’s OOD performance but also outperformed the pretrained CLIP baseline. This indicates that our fine-tuned model achieved superior generalizability, which we attribute to the integration of our custom similarity component. In contrast, RemoteCLIP showed a performance drop below the pretrained CLIP baseline on these datasets, further highlighting the advantages of our proposed approach.

% ====================================================================================
% ====================================================================================
% ====================================================================================

\subsection{Face Recognition}

In our face recognition experiments, we evaluate our model's OOD performance by testing its ability to generalize across different domains. We train on natural face images and test on cartoon and animated faces, presenting an extreme domain shift.

\subsubsection{Datasets}

Following recent work in face recognition \cite{li23bionet, huang2020curricularface, he2022enhancing}, we fine-tune our model on the MS1MV2 dataset \cite{ms1mv2}, comprising approximately 5.8 million natural facial images of 85,000 real identities. For ID evaluation, we test our tuned model on the common and standard evaluation dataset IJB-C \cite{ijbc}, composed of 3,531 real subjects with 31,334 natural images and 117,542 video frames, resembling the domain of the training data.

To assess OOD performance on extreme domain shifts, we perform our OOD evaluation on the iCartoonFace dataset \cite{icartoon_dataset}. This dataset, the largest for animated and cartoon face recognition, consists of 389,678 images of 5,013 identities. The test split is composed of 20,000 images of 2,000 identities. iCartoonFace provides a challenging testbed for evaluating the ability to recognize face identities in a domain vastly different from the training data, as can be seen in fig. \ref{fig:id_ood_examples}.

\subsubsection{Evaluation protocol}

% For the face recognition benchmarks IJB-C and iCartoonFace, we follow the standard evaluation procedures for face recognition \cite{deng2019arcface, li23bionet, huang2020curricularface, kim2020groupface}. We report our performance using the 1:1 Verification protocol, which measures the trade-off between the false accept rate (FAR) and the true accept rate (TAR) at different operating points. The TAR@FAR=$[1e-6,1e-5,1e-4]$ is a commonly reported range for IJB-C, representing the proportion of genuine matches accurately accepted at a low false accept rate in this range.

% For the iCartoonFace dataset, which represents an extreme domain shift from the natural face images used during training, we adopt a slightly modified FAR range to better examine the model's generalization capabilities. Since the domain gap between the training and test distributions is significant, we focus our evaluation on higher False Acceptance Rate (FAR) values of 1e-2 and 5e-2. These operating points allow us to assess the model's ability to recognize cartoon face identities while tolerating a higher rate of false positives, providing insights into the robustness and transferability of the learned embeddings.

For IJB-C and iCartoonFace benchmarks, we follow standard face recognition evaluation procedures \cite{deng2019arcface, li23bionet, huang2020curricularface, kim2020groupface}. We use the 1:1 Verification protocol, measuring the True Accept Rate (TAR) at different False Accept Rates (FAR). For IJB-C, we report TAR@FAR=$[1e-6,1e-5,1e-4]$. For iCartoonFace, due to the extreme domain shift, we evaluate at higher FAR values of $1e-2$ and $5e-2$ to better assess the model's generalization capabilities on cartoon face identities.

\subsubsection{Results}

We use ArcFace \cite{deng2019arcface} as our primary baseline, as it holds as one of the strongest baselines to date, and has provided the training framework widely used in face recognition models to date.
For our CLIP evaluations, we start with evaluating the pretrained CLIP without fine-tuning. Then we apply our CLIP-based implementation with contrastive learning (alg. \ref{alg:face_recognition}) with and without similarity loss. For additional comparison with ArcFace we adapt our method to use the ArcLoss (with and without similarity loss).

Table \ref{tab:face_recognition} presents our experimental results, which demonstrate several key findings. For ID, as expected and acknowledged in sec. \ref{approach_facerec}, ArcFace remains superior, given its specialization for natural face recognition.
Our main observation is the results and trend for the OOD dataset. While ArcFace struggles with this dataset, our CLIP-based method with contrastive learning and similarity loss significantly outperforms all other approaches.

Notably, the key finding is that the addition of similarity loss consistently improves OOD performance across all variants. This observation demonstrates the effectiveness of our proposed loss not only on preserving generalization capabilities that are hurt in standard fine-tuning, but also improves upon that.
ArcLoss variants show a similar trend, but with a significant drop in ID performance, suggesting that CLIP is better suited for contrastive fine-tuning.

%% file: sec/6_conclusions.tex
\section{Conclusion}\label{sec:conclusion}

In this paper, we introduced a simple but novel "similarity loss" approach to preserve the generalization capabilities of foundational models during fine-tuning for downstream tasks. Our method demonstrated significant improvements OOD performance while maintaining strong ID results across two diverse tasks: satellite imagery classification and face recognition. Notably, the trade-off between ID and OOD performance can be controlled during training through the weighting of the similarity loss, allowing for flexible adaptation to different requirements.

% An interesting future direction would be to analyze the pre-trained embeddings and potentially constrain only the components related to content rather than appearance or style.

%% file: sec/X_suppl.tex
\clearpage
\maketitlesupplementary

\section{Implementation Details}
\subsection{Image Classification}
We initialize both our fine-tuned model $f_\theta$ and the frozen baseline model $f_{\theta_0}$ with the weights of the ViT-B/32 pretrained model from OpenCLIP \cite{openclip}. We allow the entire model layers to be trained in order to adapt to the down-stream task. We used the loss function outlined in eq. \ref{eq:classification_loss}.

% We use the AdamW optimizer~\cite{adamw} with a learning rate of 5e-5 and a linear decay learning rate schedule. The batch size was set to 128 across all experiments. we experimented with values $\alpha$=[0.1,1,100,1000] and ended setting $\alpha$ to 100, as we found it the be most beneficial in terms of similarity component magnitude in comparison to the CLIP loss.

For our optimization process, we employed the AdamW optimizer \cite{adamw} with an initial learning rate of 5e-5 and implemented a linear decay learning rate schedule. We maintained a consistent batch size of 128 across all experiments. To find the right balance between our custom similarity loss and the CLIP loss, we conducted experiments with various values of the hyper-parameter $\alpha$=[0.1, 1, 100, 1000]. After careful evaluation, we determined that $\alpha$=100 yielded the most favorable results when looking at our training progress and our ID metrics, striking an optimal balance between the magnitude of the similarity component and the CLIP loss. We recommend conducting similar exploration when utilizing our solution as each data and loss can behave differently.

\subsection{Face Recognition}

We initialize both our fine-tuned model $f_\theta$ and the frozen baseline model $f_{\theta_0}$ with the weights of the ViT-B/32-laion2b pretrained model from OpenCLIP \cite{openclip}. We follow algorithm \ref{alg:face_recognition} as our training procedure.

For training, we use the AdamW optimizer~\cite{adamw} with a learning rate of 1e-5 and a linear decay learning rate schedule. The batch size is set to 256 across all experiments. We set the weight $\alpha$ for the similarity loss to 1. In addition, we found that increasing the softmax temperature to $\tau=0.1$ (in the original implementation $\tau=0.01$) improved the results.